\documentclass[runningheads]{llncs}
\pdfoutput=1
\usepackage{graphicx}
\usepackage{multirow}
\usepackage{tabularray}
\UseTblrLibrary{booktabs}
\usepackage{graphicx}
\usepackage{amsmath}
\usepackage{cite}
\usepackage{bbding}
\newcommand{\printfnsymbol}[1]{%
  \textsuperscript{\(\star\)}%
}
\begin{document}
\title{Probing Network Decisions: Capturing Uncertainties and Unveiling Vulnerabilities without Label Information}
\titlerunning{Probing Network Decisions}
%
\author{Youngju Joung\inst{1}\thanks{These authors contributed equally to this work.} \and Sehyun Lee\inst{1}\printfnsymbol{} \and Jaesik Choi\inst{1,2}\thanks{Corresponding author.}}
\authorrunning{Joung et al.}
%
\institute{Korea Advanced Institute of Science and Technology, Daejeon, Korea \\
\email{\{ojoo\_o, sehyun.lee, jaesik.choi\}@kaist.ac.kr} \and
INEEJI, Gyeonggi, Korea
}

\maketitle              
\begin{abstract}
To improve trust and transparency, it is crucial to be able to interpret the decisions of Deep Neural classifiers (DNNs). Instance-level examinations, such as attribution techniques, are commonly employed to interpret the model decisions. However, when interpreting misclassified decisions, human intervention may be required. Analyzing the attributions across each class within one instance can be particularly labor-intensive and influenced by the bias of the human interpreter. In this paper, we present a novel framework to uncover the weakness of the classifier via counterfactual examples. A prober is introduced to learn the correctness of the classifier's decision in terms of binary code - \textit{hit} or \textit{miss}. It enables the creation of the counterfactual example concerning the prober's decision. We test the performance of our prober's misclassification detection and verify its effectiveness on the image classification benchmark datasets. Furthermore, by generating counterfactuals that penetrate the prober, we demonstrate that our framework effectively identifies vulnerabilities in the target classifier without relying on label information on the MNIST dataset.

\end{abstract}

Interpretable machine learning is crucial for explaining the decisions of Deep Neural classifiers (DNNs). In particular, mission-critical systems in the real world, such as autonomous driving or AI-assisted medical diagnosis programs, should provide suitable reasons why the classifier makes such decisions. However, it is challenging to comprehend the decisions due to the complexity of these models. 

For instance-level explanation, the attribution techniques are commonly employed to highlight the influence of input features on the model decisions. By observing the gradient or activations of the classifiers, the decisions can be interpreted by calculating the saliency of input features. However, it requires human intervention to interpret the salient features since the techniques produce multiple attributions across the classes. Moreover, some cases are reported where the input attribution techniques show disagreement within the same classifier by issuing the robustness of explanations \cite{distil, RAP}

\begin{figure}[h!]
\begin{center}
\includegraphics[width=0.8\linewidth]{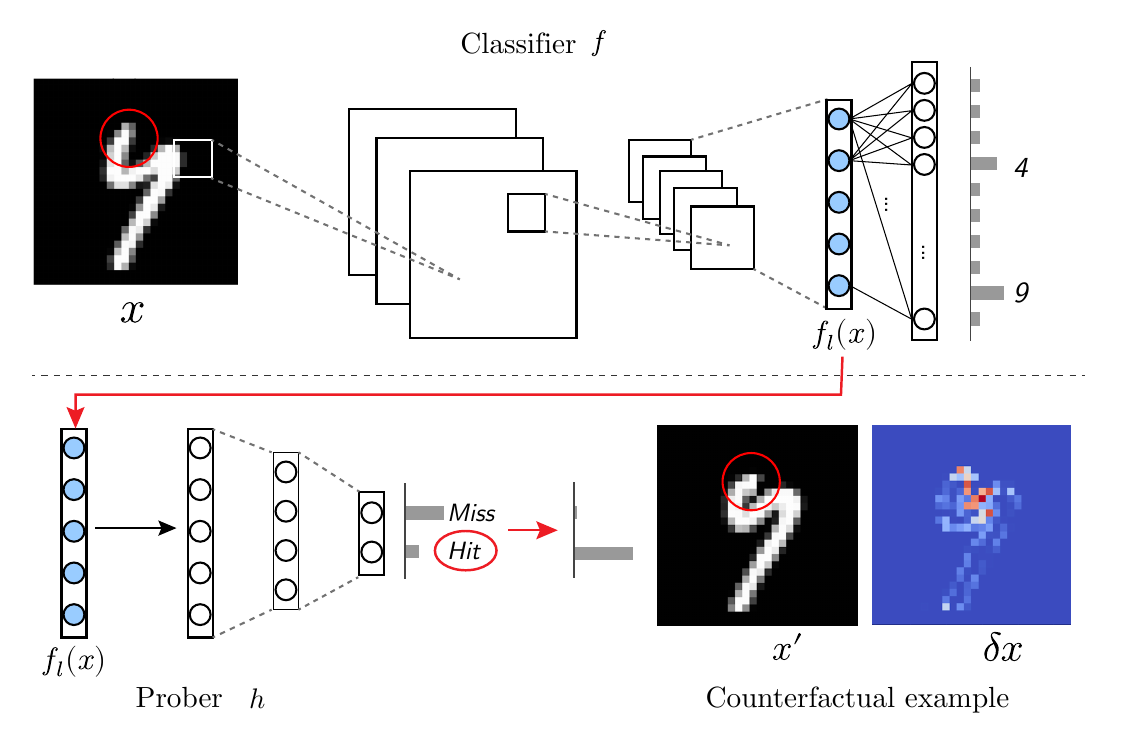}
\caption{Overview of the proposed framework to investigate the misclassified samples. In this example, the sample is classified as 9 while the true label is 4. Given the hidden representation of the layer of the instance, the prober predicts whether the classifier is \textit{hit} or \textit{miss}. Then, the counterfactual example is generated through the classifier and the prober. In this process, \textit{obstructive} features are modified, contributing to the reduction of uncertainty in the classifier's confidence.}
\label{fig:framework}
\end{center}
\end{figure}

Our goal is to reduce human interference and produce more objective explanations in interpreting the classifier. The main idea involves introducing a prober to encode the classifier's decision into binary code (\textit{hit} and \textit{miss}) so that it simplifies the number of target classes. In detail, \textit{hit} represents cases where the classifier correctly predicts the answer, while \textit{miss} indicates cases where the classifier fails to predict the correct answer. This binary encoding approach reduces the effort required for users to compare interpretations across various classes.

In this paper, we propose a simple yet novel framework to discover vulnerabilities in the classifier at the instance-level. Our contributions are as follows. First, by introducing a shallow Feed Forward Network (FFN) called a prober, we conduct detection for misclassified samples. Additionally, we construct a \textit{Hit}-\textit{Miss} Dataset for the experiment and explore the evidence behind how the prober operates effectively. Second, through analyzing the prober, we indirectly investigate the classifier to identify \textit{obstructive} features that confuse the classifier. Since our method employs a binary encoding approach, it is enable to access potential labels even in opaque scenarios where the true label is unknown. Moreover, the advantage of our framework demonstrates the possibility of facilitating auto-correction capabilities for the classifier.

\section{Related Work}

\textbf{Prober}.
Probers are widely applied for testing specific properties of the internal representations within a neural classifier. Exploring the internal workings of a neural classifier is typically challenging, so employing a relatively simple prober for indirect analysis is preferred.~\cite{ProberNLPICLR2017,ProberNLPACL2018,ProberNLPEACL2021}. In the case of Natural Language Processing (NLP), probing techniques are extensively utilized to evaluate knowledge of Language Models (LM). For example, detecting true-false relationships can be employed to validate the model's ability to handle hallucinations~\cite{InternalStateLLM2023}. Predicting verb tense is used to assess the model's understanding of sentence structure~\cite{ProberNLPNAACL2018}. To prevent additional biases and ensure the simplicity of the prober, a simple feedforward classifier is typically employed to measure probing accuracy. \\

\textbf{Misclassification Detection}. Misclassification detection aims to identify the misclassified samples from the given classifier to enhance the reliability of confidence estimation. It is closely related to out-of-distribution (OOD) detection, but misclassification is not constrained to whether the sample comes from in-distribution or out-of-distribution explicitly. A baseline of misclassification detection is using the maximum softmax probability \cite{baseline}. OpenMix \cite{OpenMix2023CVPR} proposed utilizing the outlier exposure method \cite{OutlierExposure} for more reliable misclassification detection. However, extra actions are still required to interpret the misclassified samples. \\

\textbf{Counterfactual Explanations}. 
The counterfactual-based explanation \cite{kusner2018counterfactual, DBLP:journals/corr/abs-2010-10596} method aims to explain the causal relationship between input and the model's final decision. In the classification task, to understand why the input could not be classified into another class, a situation opposite to what is actually observed is assumed. As counterfactual examples aim to provide semantically different examples compared to adversarial examples, a generative model is employed to generate more semantically meaningful counterfactual examples \cite{Diffeomorhpic}. Specifically, they propose the Approximated Diffeomorphic Counterfactuals (ADC) approach, utilizing normalizing flow to find optimal values in the latent space of the generator along the data manifold. We embrace the ADC method for discovering and analyzing vulnerabilities in the classifier.

\section{Methodology}
In this section, we first suggest the proposed framework to provide an explanation through counterfactual examples representing a likely \textit{hit} decision from the perspective of the classifier as illustrated in Figure~\ref{fig:framework}. Then, we present the following three core elements comprising this framework including a \textit{Hit}-\textit{Miss} datasets, a prober, and a counterfactual generator. 

\subsection{Probing Framework}
In this subsection, we present an overview of the probing framework along with relevant notations. The framework is designed to work with a classifier \(f\) of interest, initially trained for a specific task \(x \mapsto \hat{y}\), where \(x\) represents the input and \(\hat{y}\) is the prediction. The primary focus of this paper is on the image classification task, but it should be noted that the framework is versatile and adaptable to diverse domains and tasks.

The classifier evaluates an annotated dataset \(\mathcal{D}=\{x^{(i)}, y^{(i)}\}\) and generates the intermediate representation of \(x\) at the layer \(l\), denoted as \(f_l(x)\). To encode the classifier’s decision into binary code, a probing label \(o\) is scored to indicate whether the classifier predicts the class correctly or incorrectly. Following this, a prober \(h : f_l(x) \mapsto \hat{o} \) is trained on the \textit{Hit}-\textit{Miss} dataset \(\mathcal{D}_P = \{f_l(x)^{(i)}, o^{(i)}\} \), which consists of representations and the corresponding probing label. The prober allows for the indirect analysis of the \textit{obstructive} feature that straightforwardly hindered the prediction of the classifier.

\subsection{The \textit{Hit}-\textit{Miss} Dataset}
In the context of our work, we leverage a \textit{Hit}-\textit{Miss} dataset  \(\mathcal{D}_P = \{f_l(x)^{(i)}, o^{(i)}\} \) encoded in binary form \(o=\{0,1\}\), enabling us to frame the problem as binary classification. A positive class represents the prediction of the classifier is the same as the true class - \textit{hit}. In contrast, a negative class indicates the prediction is different - \textit{miss}.

We assume the decisions of the classifier are reliable, believing that the classifier correctly captures the meaningful features. However, this assumption introduces an imbalance in the \textit{Hit}-\textit{Miss} dataset concerning probing labels. A substantial difference in counts between \textit{hit} and \textit{miss} instances is observed in Table~\ref{tab:classifier}. This imbalance poses a challenge for the prober, as it may be prone to consistently indicating that the classifier is always \textit{hit}. Strategies to address this issue will be discussed in the following subsection.

\subsection{Prober}
To capture the hidden features of the target classifier without becoming overly complex and without introducing additional biases, a simple Feed Forward Network (FFN) is chosen. This involves using three fully connected layers with ReLU activations. The prober takes the hidden representations of a specific layer in the target classifier as input and produces an output indicating whether the classifier classifies the given sample correctly or not. Due to the significant imbalance ratio between the \textit{hit} and \textit{miss} labels, two strategies are employed for mitigation: adjustment of loss weight and application of label-smoothing regularization \cite{label_smoothing}.
The cross-entropy objective is expressed as:
\begin{gather}
q(k) = o_k(1-\alpha)+\frac{\alpha}{K} \\
\mathcal{L} = - \sum_{k=1}  q(k) \log{p(k)} + w  (1-q(k)) \log{(1-p(k))}
\end{gather}
where \(K\) represents the number of classes and \(q(k)\) is the likelihood the model assigned to the \(k\)-th class. Here, \(K=2\), because the prober predicts whether it is \textit{hit} or \textit{miss}. In the case of \textit{hit} label that is represented as one hot vector, it is transformed from \([0, 1]\) to \([\frac{\alpha}{2}, 1-\frac{\alpha}{2}]\) given a label smoothing parameter \(\alpha\). To mitigate the imbalance of labels, the weight \(w\) is assigned to the \textit{miss} class. These adjustments help alleviate the risk of overfitting, promote generalization, and prevent excessive model confidence~\cite{LabelSmoothing}. It is empirically demonstrated how the prober shows the misclassification performance on the \textit{Hit}-\textit{Miss} dataset in section~\ref{sec:how}.

\subsection{Counterfactual Generator}
The prober evaluates the likelihood of misclassification by manipulating the internal information of the classifier. Consequently, interpreting the prober facilitates an indirect examination of the classifier's behavior. Traditional Explainable Artificial Intelligence (XAI) methods predominantly analyze models under the assumption of knowing the target label. However, an explanation is often required for unlabeled data in the real world. Hence, we specifically consider scenarios where the true label remains unknown. 

We then aim to generate Approximated Diffeomorphic Counterfactuals \cite{Diffeomorhpic} for the prober's \textit{hit} score (\(ADC_{hit}\)) to analyze the conditions under which the classifier approaches the actual answer. Given the original image \(x\), we perform gradient ascent on the \textit{hit} logit of the prober and the optimal value \(z^*\) is explored in the latent space of the generator \(g\). The optimization process for finding the optimal value \(z^*\) and generating \(ADC_{hit}\) is as follows.

\begin{gather}
z = g^{-1}(x) \\[5pt]
z^{(i+1)} = z^{(i)} + \lambda \frac{\partial((h \circ f_l) \circ g)_{hit}}{\partial z} (z^{(i)}) \\[5pt]
ADC_{hit}(x) = g(z^*)
\end{gather}

Where \(f\), \(g\), \(h\), \(\lambda\) and \(l\) denote the classifier, generative model, prober, step size, and the target layer to extract a hidden representation of the classifier, respectively. In addition, pre-trained RealNVP \cite{dinh2017density, Diffeomorhpic}  is employed as a generative model. In this study, our focus is on comprehending the limitations of the classifier rather than emphasizing correctly predicted samples. Therefore, the primary analysis set comprises samples predicted as \textit{miss} by the prober. (The results can be found in Section~\ref{sec:cf}.)

\section{Experiments}

\subsection{Experimental Setup}
\subsubsection{Datasets}
We employ four different benchmark datasets that encompass image classification task. MNIST \cite{MNIST} is a dataset of hand-written digits, consisting of grayscale images with 10 classes and a total of 60k images, characterized by low resolution. Fashion-MNIST (F-MNIST) \cite{FashionMNIST} is similar to MNIST, but the images contain 10 different types of clothing, making it slightly more complex than MNIST. CIFAR-10 \cite{Krizhevsky2009LearningML} consists of 60k color images distributed across 10 classes. ImageNette \cite{DBLP:journals/corr/abs-2002-04688} is a subset of the ImageNet \cite{5206848} dataset comprising 10 easily classified classes, representing almost 10\% of the original dataset. This allows testing performance on the high-resolution image (we use 128px).

\begin{table}[h]
\centering
\caption{Comparison of image classification accuracy of image classification benchmark datasets: MNIST, F-MNIST, CIFAR10, and ImageNette.}
\label{tab:classifier}
\begin{tblr}{
    colspec = {llcccc},
    cell{1}{3} = {c=2}{c},
    cell{1}{5} = {c=2}{c},
    row{1}     = {font=\bfseries},
}
\toprule
        &           & Train     &       & Test      &       \\
\cmidrule[lr]{3-4} \cmidrule[lr=-0.4]{5-6}
\textbf{Dataset} & \textbf{Classifier} & \textbf{Top-1} & \textbf{Top-5} & \textbf{Top-1} & \textbf{Top-5}  \\ \midrule
MNIST    & CNN       & 98.60       & 99.99       & 98.50       & 99.97      \\
F-MNIST  & CNN       & 98.03       & 99.99       & 92.86       & 99.87      \\
CIFAR10  & ResNet18  & 91.07       & 99.83       & 79.63       & 98.75    \\
ImageNette & XResNet50     & 86.63       & 99.00       & 83.92      & 98.29   \\
\bottomrule
\end{tblr}
\end{table}

\subsubsection{Model architecture}
The model architectures for each dataset were selected based on established precedents. For MNIST and F-MNIST, we use a CNN with four blocks, each comprising batch normalization, non-linear ReLU activation, max-pooling, and dropout, following the architecture outlined in \cite{Diffeomorhpic}. For CIFAR-10, we adapt a ResNet18 architecture by adjusting the kernel size of the first convolutional layer to 3 and replacing max-pooling to be an identity function in accordance with \cite{detectors}. In the case of ImageNette, we use XResNet-50 \cite{fastai} architecture. Hidden representations are extracted just before the final fully-connected layer, resulting in dimensions of 256, 256, 256, and 2048 for the respective datasets. The prober utilizes a simple feedforward classifier with three fully connected layers and non-linear ReLU activation in all models.

\subsubsection{Training details}
During the training of classifiers, we adhered to the conventional train/test dataset split with standard cross-entropy loss. When a classifier is capable of capturing the proper features, it is valuable to utilize the counterfactuals because it exhibits sufficient discriminative power. Under this assumption, the classifiers achieve high classification accuracy, resulting in an imbalance between the number of \textit{hit} and \textit{miss} classifications in the dataset, as detailed in Table~\ref{tab:classifier}. To ensure the prober does not express excessive confidence in \textit{hit} predictions, we applied label smoothing regularization with a coefficient of 0.2 and assigned a weight of 2 to the \textit{miss} label empirically.

\begin{table}[t!]
\centering
\caption{Comparison of misclassification detection performance on benchmark datasets - MNIST, FMNIST, CIFAR10, and ImageNette. The performance are reported with \textbf{AUPR (\(\uparrow\)), AUROC (\(\uparrow\)), FPR95 (\(\downarrow\))} and \textbf{Accuracy (\(\uparrow\))}. AUPR is calculated by treating the \textit{miss} class as positive. All metrics are presented as percentages. Imbalance ratio (IR) indicates the proportion of the instances in the \textit{hit} class to the number of instances in the \textit{miss} class based on Top-1 accuracy. The larger the value of IR, the greater the extent of imbalance.}
\label{tab:prober}
\begin{tblr}{
    width      = \textwidth,
    colspec    = {lcccccc},
    row{1}     = {font=\bfseries},
}
\toprule 
Dataset & IR\textsubscript{train}/IR\textsubscript{test} & Prober & AUPR  & AUROC & FPR95 & ACC  \\ 
\midrule
MNIST & 70.8/51.3 & 256-128-64 & 39.98 & 98.37 & 3.57  & 98.60 \\
F-MNIST & 37.8/13.0 & 256-256-64 & 48.29 & 86.39 & 45.45 & 92.30 \\
CIFAR10 & 8.54/3.74 & 256-256-64 & 46.66 & 84.17 & 69.18 & 84.10 \\
ImageNette & 6.03/6.54 & 2048-512-64 & 51.44 & 84.24 & 59.50 & 84.59 \\  
\bottomrule
\end{tblr}
\end{table}

\subsection{Results of the Misclassification Detection}
We initiate the evaluation of the prober's performance by assessing its ability to accurately classify misclassified samples across diverse benchmark datasets, including MNIST, F-MNIST, CIFAR10, and ImageNette. The evaluation employs essential metrics, including Area Under the Precision-Recall curve (AUPR), Area under the Receiver Operating Characteristic curve (AUROC), False Positive Rate at 95\% True Positive Rate (FPR95), and Test Accuracy (ACC) \cite{AURC, AUROC, OpenMix2023CVPR}.

AUPR quantifies the model's performance across various levels of precision and recall, which is especially useful for an imbalanced dataset. AUROC represents the relationship between True Positive Rate (TPR) and False Positive Rate (FPR), offering the classifier's discriminatory ability. FPR95, the False Positive Rate at 95\% TPR, indicates the probability of misclassified examples being predicted as \textit{hit} when the TPR reaches 95\%. Test Accuracy (ACC) serves as a fundamental metric to assess overall classifier performance. Collectively, these metrics facilitate a comprehensive evaluation of the prober's achievement in identifying misclassified samples across diverse datasets.

Table~\ref{tab:prober} presents the performance of the prober, empirically demonstrating the ability to predict the correctness of classifier predictions based on hidden representation. Overall, the prober achieved high accuracy and AUROC across all datasets. The significance lies in its competent success, even on high-resolution image dataset such as ImageNette. Notably, for MNIST, the prober shows exceptional performance, particularly in terms of AURC and FPR95. Due to the high imbalance, it was challenging to learn the proper relationship between hidden representation and \textit{Hit}-\textit{Miss} label. With a small number of \textit{miss} instances in the test dataset, since the prober has high confidence in \textit{hit}, it causes a poor FPR95. Relatively, AURC captures the precision-recall trade-off over the thresholds, showing a different aspect than FPR95. There is much room for accurately improving the prober to learn the \textit{Hit}-\textit{Miss} dataset. Throughout the remainder of the study, we analyze the MNIST dataset and use the prober that shows the best performance among datasets.

\begin{figure}[t!]
\centering
\includegraphics[width=\linewidth]{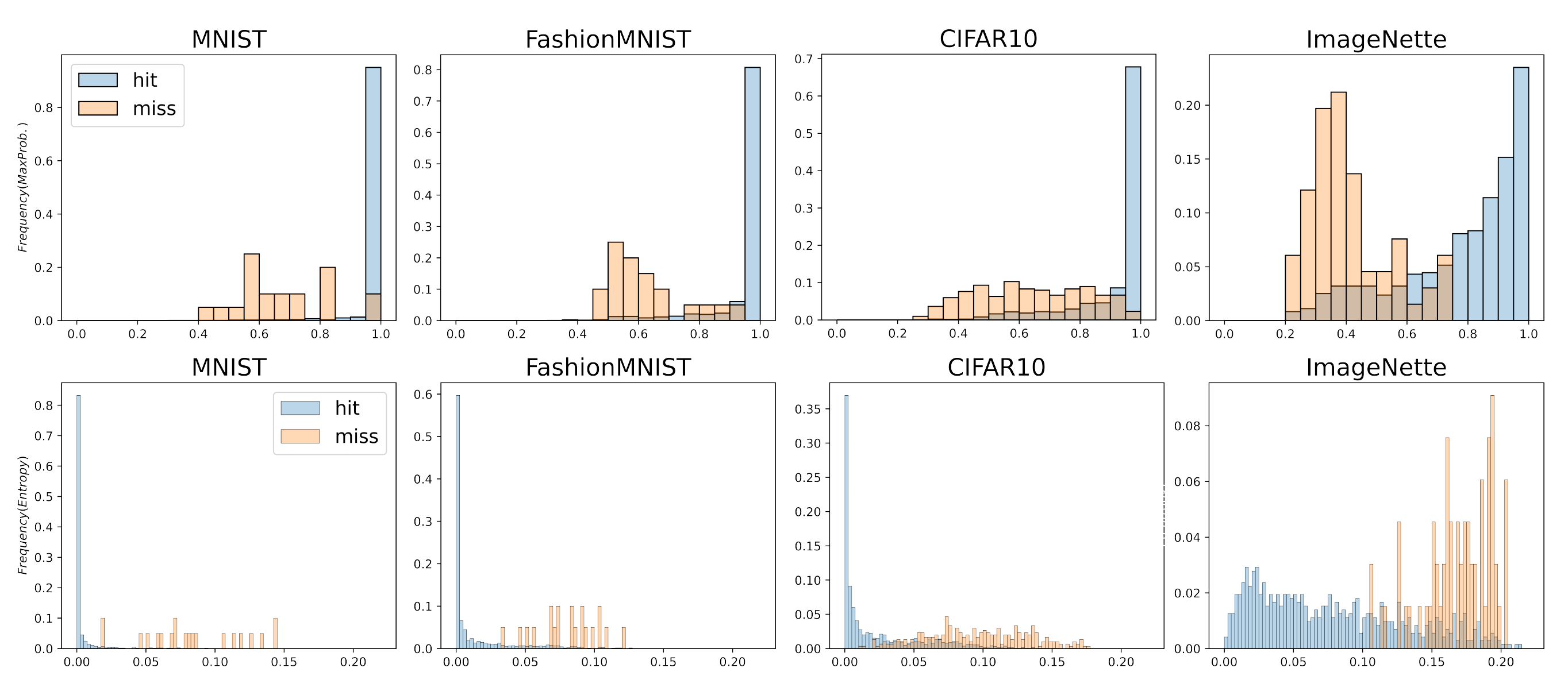}
\caption{Maximum probability (first row) and entropy of probability (second row) of the classifier.} 
\label{fig:max_ent}
\end{figure}

\subsection{How the Prober Works}\label{sec:how}
As mentioned above, despite employing a compact FFN model as the prober, It adeptly anticipates the actions of the classifier for a variety of datasets. Then, how does the prober possess the foresight to discern whether the classifier is lying or not? In addressing this question, we hypothesize that the prober captures information about \textit{uncertainty} within the hidden representation of the classifier.

To validate the hypothesis, we partition the data samples into two groups based on the prober's predictions (\textit{hit} and \textit{miss}). Subsequently, we observe the classifier's behavior when classifying these samples. Figure~\ref{fig:max_ent} illustrates the frequency of (1) the probability of the classifier's final prediction and (2) the entropy of probability for instances categorized into the \textit{hit} and \textit{miss} groups, respectively. It is evident that the \textit{hit} group exhibits a higher maximum probability and lower probability entropy compared to the \textit{miss} group. More closely, we conduct a statistical test to determine whether there is a significant difference in the medians between the two groups. The hypothesis for testing is formulated as follows where \(m^{prob}_{A}\), \(m^{ent}_{A}\) represents the medians of the maximum probability and probability entropy of group \(A\) respectively.

\begin{gather}
\label{eq:hypotheses}
\begin{cases}
H_0^{prob}: m^{prob}_{hit} \leq m^{prob}_{miss}\\[3pt]
H_1^{prob}: m^{prob}_{hit} > m^{prob}_{miss}
\end{cases} , \quad
\begin{cases}
H_0^{ent}: m^{ent}_{hit} \geq m^{ent}_{miss}\\[3pt]
H_1^{ent}: m^{ent}_{hit} < m^{ent}_{miss}
\end{cases}
\end{gather}

\begin{figure}[t!]
\begin{center}
\includegraphics[width=0.85\linewidth]{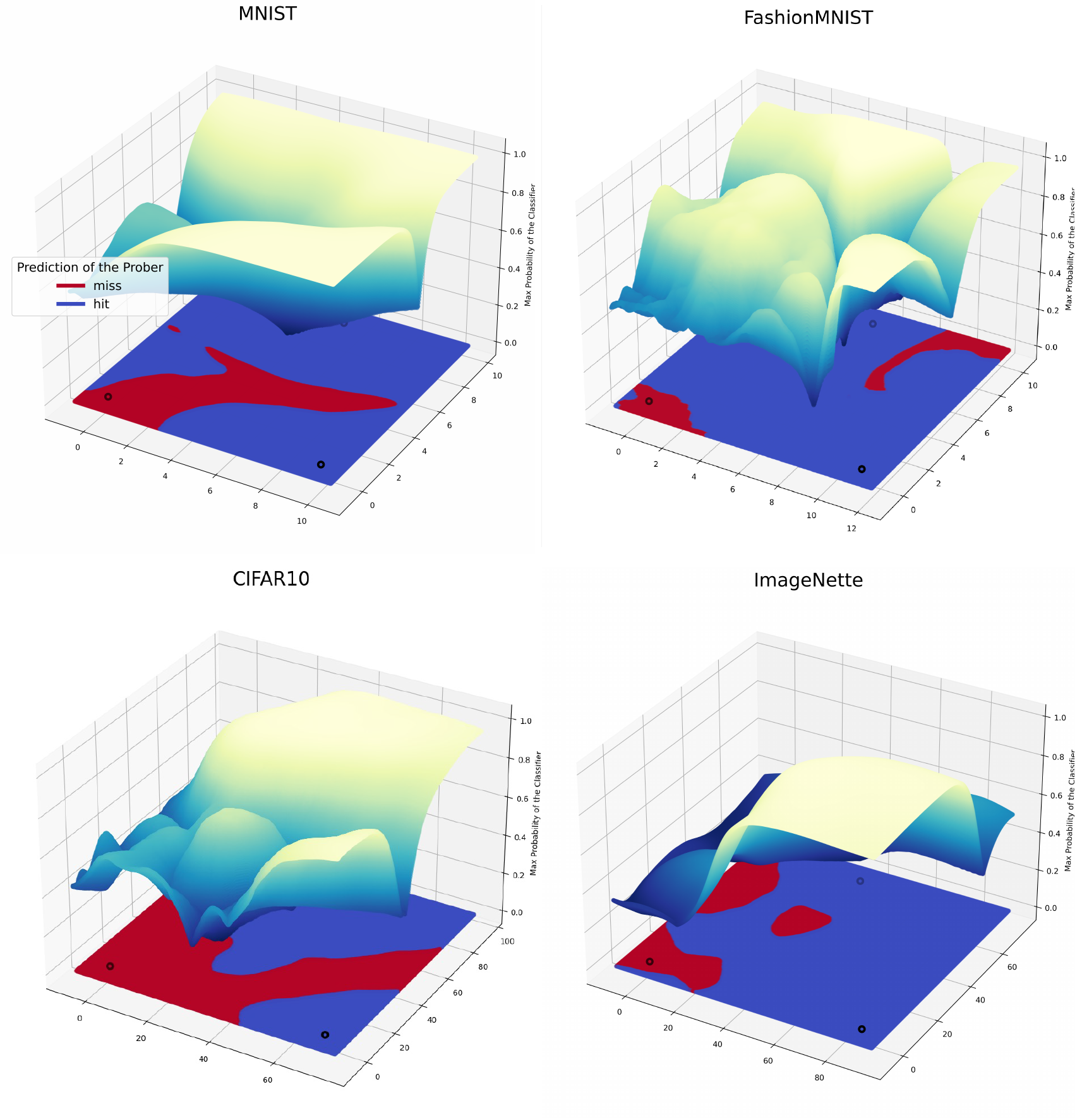}
\caption{The action of the prober according to the maximum probability of the classifier. The \(xy\)-plane displays a 2D plane formed by three selected points from the dataset. At the bottom, for samples lying on the plane, the predicted action of the classifier (\textit{miss} or \textit{hit}) by the prober is denoted in red or blue. The \(z\)-axis represents the probability of the prediction when the sample is fed into the classifier. It implies that the prober tends to output \textit{hit} for samples where the classifier is confident and \textit{miss} when the classifier lacks confidence.} 
\label{fig:3dplot}
\end{center}
\end{figure}

In most cases, since the normality assumption is not satisfied, we follow the non-parametric approach known as the Mann-Whitney U-test \cite{article}. As indicated in Table~\ref{tab:test}, the null hypotheses are rejected at the 5\% significance level, so that the alternative hypotheses are adopted. It suggests that samples within the \textit{hit} group exhibit significantly higher maximum probability and lower probability entropy on the classifier compared to the \textit{miss} group. This supports our conjecture that the prober would detect \textit{uncertainty} from the hidden representation of the classifier. Indeed, upon visualizing the classifier's hidden space, the prober tends to output \textit{hit} for samples where the classifier is confident and \textit{miss} where the classifier lacks confidence (See Figure~\ref{fig:3dplot}).

\begin{table}[]
\centering
\caption{Results of the Mann-Whitney U-test. Non-parametric approach is employed to assess significant differences between the two groups. The hypothesis is formulated as in equation (\ref{eq:hypotheses}). The alternative hypotheses are adopted under the 5\% significance level across all cases. In other words, it can be confirmed that the classifier exhibits low uncertainty for samples within the \textit{hit} group and high uncertainty for samples within the \textit{miss} group. The experiment was conducted on both the train and test sets of the prober. (* \(p<0.05\))}
\label{tab:test}
\begin{tblr}{
    colspec      = {llcccc},
    cell{1}{3}   = {c=2}{c},
    cell{1}{5}   = {c=2}{c},
    row{1-2}     = {font=\bfseries},
    column{3-6}  = {}{c},
    cell{3}{1}   = {r=2}{valign = m},
    cell{5}{1}   = {r=2}{valign = m},
    cell{7}{1}   = {r=2}{valign = m},
    cell{9}{1}   = {r=2}{valign = m},
}
\toprule
      &          & Train &      & Test  &    \\
\cmidrule[lr]{3-4} \cmidrule[lr=-0.4]{5-6}
Dataset & Value  & U     & p-value  & U     & p-value \\
\midrule
MNIST & Max-Prob & 24878068.5*     & 3.52e-253       & 57.5*    & 4.31e-14 \\
      & Entropy & 620721.0*    & 3.52e-253      & 665.0*    & 5.64e-9 \\
\midrule
F-MNIST & Max-Prob & 20804015.0*    & 4.56e-213       & 37998.0*    & 7.12e-13 \\
      & Entropy & 523957.5*    & 1.33e-211       & 1742.0*    & 1.05e-12\\
\midrule
CIFAR10 & Max-Prob & 215370525.5*    & 0.0       & 464464.0*    & 1.84e-113 \\
      & Entropy & 917.51*    & 0.0       & 39471.0*   & 1.49e-121 \\
\midrule
ImageNette & Max-Prob & 17172817.0*    & 0.0       & 43285.0*   & 6.76e-29 \\
      & Entropy & 2031060.0*    & 0.0       & 3970.0*    & 1.90e-29 \\
\bottomrule
\end{tblr}
\end{table}

\subsection{Uncovering Weaknesses via Counterfactual-based Analysis}\label{sec:cf}

\begin{figure}[t!]
\begin{center}
\includegraphics[width=0.73\linewidth]{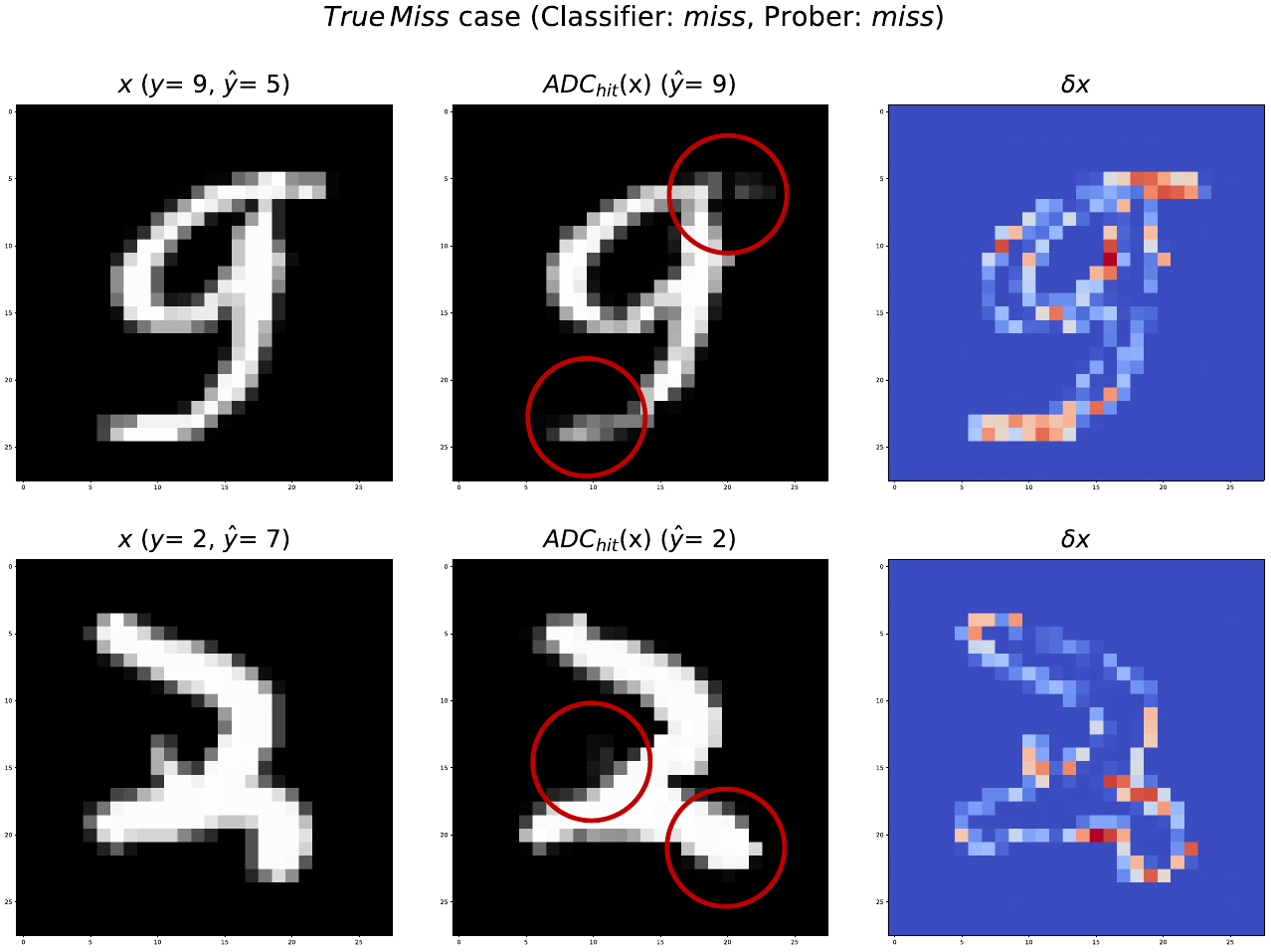}
\caption{Counterfactual examples \(ADC_{hit}(x)\) generated for the \textit{True Miss} cases where both the classifier fails to classify accurately, and the prober predicts as \textit{miss}. \(y\) and \(\hat{y}\) denote the label and the prediction of the classifier, respectively. Despite lacking information about the true label, the prober identifies vulnerabilities in the classifier for each sample \(x\). With this framework, we obtain examples close to the correct answer by modifying the regions indicated in red, corresponding to \(\delta x\).}
\label{fig:TM}
\end{center}
\end{figure}

\begin{figure}[t!]
\begin{center}
\includegraphics[width=0.73\linewidth]{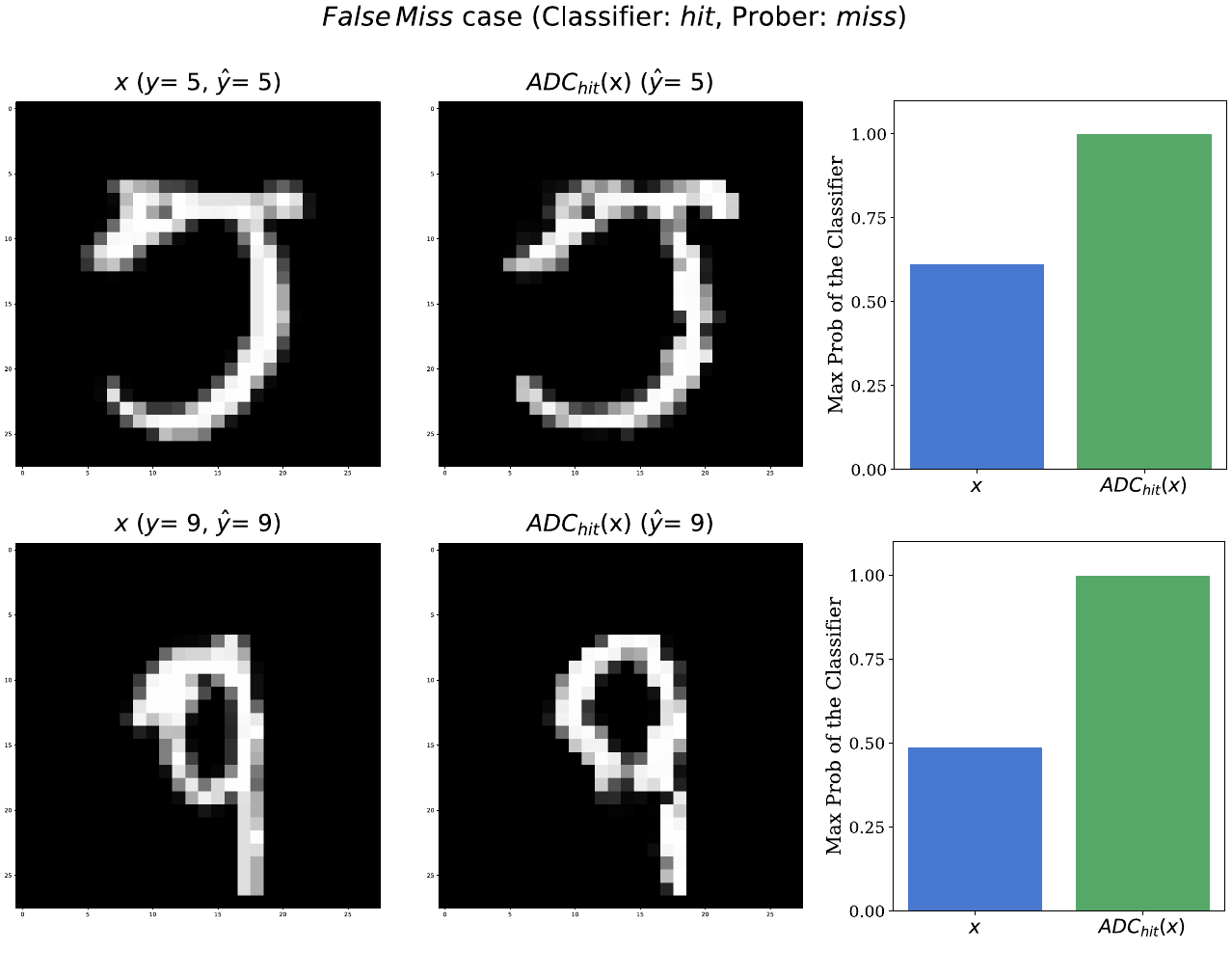}
\caption{Counterfactual examples \(ADC_{hit}(x)\) generated for the \textit{False Miss} cases where the classifier correctly classifies, but the prober predicts as \textit{miss}. \(y\) and \(\hat{y}\) denote the label and the prediction of the classifier, respectively. Even though the original image \(x\) is already correctly classified, \(ADC_{hit}(x)\) is generated in a direction aimed at reducing uncertainty from the perspective of the classifier.} 
\label{fig:FM}
\end{center}
\end{figure}

\begin{table}[h!]
\centering
\caption{Re-classification experiments for \(ADC_{hit}\) of MNIST: changes in the classifier's accuracy and maximum probability. The accuracy is calculated for the true label, and the variation of the maximum probability is averaged across all target samples. The classifier demonstrates high accuracy and confidence for \(ADC_{hit}\) generated in an unsupervised manner. This implies that the prober adeptly captures the weakness of the classifier, without relying on label information and regardless of whether the classifier correctly predicts the actual label or not.}
\label{tab:reclassification}
\setlength{\tabcolsep}{6pt} 
\renewcommand{\arraystretch}{1.2} 
\begin{tabular}{@{}cccc@{}}
\toprule
                                                  & \multicolumn{3}{c}{\textbf{Prediction of the Prober}}       \\ \midrule
                                                  & \textbf{\textit{Miss}} & \textbf{\textit{True Miss}} & \textbf{\textit{False Miss}} \\ \cmidrule(l){2-4} 
\multicolumn{1}{r}{\textbf{Accuracy (\%)}}        & 25 → 90        & 0 → 86.67           & 100 → 100            \\
\multicolumn{1}{r}{\textbf{\(\Delta\) Max Probability (\%)}} & +29.21        & +29.21             & +33.40              \\ \bottomrule
\end{tabular}
\end{table}

Figure~\ref{fig:TM} and \ref{fig:FM} illustrate the counterfactual examples of performing gradient ascent in the latent space for the \textit{hit}-logit of the prober, called \(ADC_{hit}\), on the MNIST dataset. The analysis is conducted only on samples that the prober classifies as \textit{miss} in the test set. Among them, we categorize cases where the classifier accurately predicts the true label as \textit{True Miss} and cases where the classifier fails to predict the true label as \textit{False Miss}. As mentioned earlier, we assume an opaque situation regarding the labels. Therefore, to generalize over all \textit{miss} instances independent of the classifier's behavior, we divide the analysis into two cases. As depicted in Figure~\ref{fig:TM}, for the \textit{True Miss} samples, \(ADC_{hit}\) provides examples that closely align with the potential label as viewed from the prober's perspective based on the original image \(x\). It is noteworthy that the prober operates without any knowledge of the true labels. Despite this lack of information, the prober can manipulate instances by emphasizing or removing certain regions \(\delta x\) that might confuse the classifier. In the case of \textit{False Miss} where the classifier has already correctly predicted the answer, as presented in Figure~\ref{fig:FM}, it can be inferred that counterfactual examples are generated in a direction that reduces the uncertainty of the classifier.

To clarify, we experiment to re-classify using the newly generated \(ADC_{hit}\) examples. Table \ref{tab:reclassification} reports the changes in classifier accuracy for true labels and the maximum probability for the final prediction when replacing original images with \(ADC_{hit}\) samples. The results indicate that, for the \textit{True Miss} samples, \(ADC_{hit}\) leads to an approximately 86.67\% improvement in classifier accuracy. Regarding \textit{False Miss} cases, where the original answers were correctly predicted, it's essential to note that the starting accuracy is 100\%. However, with an increase of around 33.40\% in the maximum probability for the final decision, it is concluded that \(ADC_{hit}\) can be an example that complements the classifier's weaknesses. As a result, the introduction of the prober holds significance in that it allows for the identification of vulnerabilities in the classifier at the instance-level. Indeed, it is advantageous in providing interpretations in all cases, even without information about the true label and regardless of the predictions made by the classifier.

\section{Conclusion}
In this paper, we presented a framework for addressing the weakness of the classifier by providing counterfactual examples. The prober was trained to predict whether the classifier's decision is correct (\textit{hit}) or incorrect (\textit{miss}) based on the hidden representation of the classifier. The underlying hypothesis is that the prober can capture the confidence of the classifier's predictions by inferring decision probabilities. We have successfully validated that the prober captures the uncertainty in the classifier's decision by investigating the maximum and entropy of probability. Moreover, we generated semantically meaningful counterfactual examples to demonstrate how the input should appear if the prober expects the classifier to output the correct answer. It edited the indiscriminative features that obstruct the classifier's decision without access to the true label. Future developments can include the improvement of the performance of the prober and the counterfactual explanations in more complicated datasets. Another possible future direction is expanding to auto-correction and utilizing our framework to address model-level interpretation.

\subsubsection{Acknowledgement} 
This work was conducted by Center for Applied Research in Artificial Intelligence (CARAI) grant funded by DAPA and ADD (UD230017TD) and partly supported by Institute of Information \& Communications Technology Planning \& Evaluation (IITP) grant funded by the Korea government (MSIT) (No. 2022-0-00984, Development of Artificial Intelligence Technology for Personalized Plug-and-Play Explanation and Verification of Explanation; No. 2019-0-00075, Artificial Intelligence Graduate School Program (KAIST)).

\bibliographystyle{splncs04}
\bibliography{main}
\end{document}